\documentclass[a4paper,fleqn]{cas-dc}

\usepackage[numbers]{natbib}
\usepackage{algorithm}
\usepackage{algpseudocode}
\usepackage{enumitem}
\newcolumntype{L}{>{\color{black}}l} 

\newcommand{\blue}[1]{\textcolor{black}{#1}}
\def\tsc#1{\csdef{#1}{\textsc{\lowercase{#1}}\xspace}}
\tsc{WGM}
\tsc{QE}

\begin{document}
\let\WriteBookmarks\relax
\def\floatpagepagefraction{1}
\def\textpagefraction{.001}

\shorttitle{}    

\shortauthors{Kamel et~al.}  


\title [mode = title]{Improving action segmentation via explicit similarity measurement} 



%

\author[1]{Kamel Aouaidjia}[orcid=0000-0001-6286-9527]



\ead{kamel@henu.edu.cn}


\credit{Conceptualization, Experiments, Writing original draft}


\affiliation[1]{organization={School of Computer
and Information Engineering, Henan University},
            city={Kaifeng},
            postcode={475001}, 
            state={Henan},
            country={China}}


\author[1]{Wenhao Zhang}


\ead{zxs77889@henu.edu.cn}


\credit{Implementation, Experiments}

\author[1]{Aofan Li}


\ead{henulaf@henu.edu.cn}


\credit{Implementation, Experiments}

\author[1]{Chongsheng Zhang}


\ead{cszhang@ieee.org}
\cormark[1]

\credit{Writing, Reviewing and Editing}

\cortext[1]{Corresponding author.
E-mail: cszhang@ieee.org  (C. Zhang)}



\begin{abstract}
Existing supervised action segmentation methods depend on the quality of frame-wise classification using attention mechanisms or temporal convolutions to capture temporal dependencies. Even boundary detection-based methods primarily depend on the accuracy of an initial frame-wise classification, which can overlook precise identification of segments and boundaries in case of low-quality prediction. To address this problem, this paper proposes ASESM (Action Segmentation via Explicit Similarity Measurement) to enhance the segmentation accuracy by incorporating explicit similarity evaluation across frames and predictions. Our supervised learning architecture uses frame-level multi-resolution features as input to multiple Transformer encoders. The resulting multiple frame-wise predictions are used for similarity voting to obtain high quality initial prediction. We apply a newly proposed boundary correction algorithm that operates based on feature similarity between consecutive frames to adjust the boundary locations iteratively through the learning process. The corrected prediction is then further refined through multiple stages of temporal convolutions.  As post-processing,  we optionally apply boundary correction again followed by a segment smoothing method that removes outlier classes within segments using similarity measurement between consecutive predictions. Additionally, we propose a fully unsupervised boundary detection-correction algorithm that identifies segment boundaries based solely on feature similarity without any training. Experiments on 50Salads, GTEA, and Breakfast datasets show the effectiveness of both the supervised and unsupervised algorithms. Code and models are made available on Github\footnotemark.

\nocite{*}
\end{abstract}








\maketitle


\section{Introduction}\label{section1}
Action segmentation in videos is essential for many applications such as surveillance, sports analysis, and medical assistance \cite{karbalaie2022event, yang2023abnormal, wu2022survey, liu2018deep}. It involves the semantic interpretation of distinct events and their temporal location in an untrimmed video. The challenge lies in distinguishing between different actions that occur within the same background, especially when the differences between consecutive actions are minimal, such as with the actions `cutting tomato` and `placing tomato into a bowl`, where both actions have the same background and objects such as hands and tomatoes. The only difference is using the bowl instead of the knife in the scene. These subtle differences require an efficient method that captures local detail changes as well as global scene variations through learning discriminative frame-level representations and capturing similar patterns of the same action across frames.

The earliest action segmentation techniques relied on modeling frame sequences using Hidden Markov Models (HMM), Gaussian Mixture Model (GMM), or a combination of both \cite{kuehne2016end, kuehne2017weakly, sener2018unsupervised}. With the emergence of deep learning,  Recurrent Neural Networks (RNNs) with its variants  are widely used for modeling action sequences, including  unidirectional RNN, bi-directional RNN, Gated Recurrent Units (GRUs) and LSTM  \cite{richard2017weakly, huang2020improving, singh2016multi}. However, they are unable to capture long-term dependencies in longer videos. Recent methods tackle the problem by capturing temporal dependencies in the sequence using self-attention mechanism in Transformer \cite{yi2021asformer,behrmann2022unified,zhu2020deformable} or temporal convolution for both feature extraction and prediction refinement through multiple
stages \cite{lea2017temporal, lei2018temporal, singhania2023c2f, farha2019ms,li2020ms}, and they showed high performance by just focusing on frame-wise prediction, where the segmentation is the result of accurate classification of consecutive frames. Over-segmentation methods focus on localizing and correcting boundaries in the sequence \cite{wang2020boundary,ishikawa2021alleviating,ding2022leveraging,ding2018weakly}.

Existing methods have the following limitations: i) Refinement-based approaches \cite{yi2021asformer, farha2019ms, li2020ms, bahrami2023much} that rely on frame-wise prediction, heavily depend on the classification accuracy of individual frames, meaning that errors in frame-wise prediction can lead to inaccurate segmentation, especially at segments boundaries. Over-segmentation methods attempt to alleviate this ambiguity by incorporating boundary detection. ii) Although methods such as \cite{ishikawa2021alleviating, wang2020boundary} introduce an additional branch to predict frame-wise boundary probabilities \cite{ishikawa2021alleviating} or the start and the end of segments \cite{wang2020boundary}, the boundary detection is also based on model predictions, lacking a mechanism to verify the accuracy of boundary location correctness explicitly.  ii) Boundary correction methods depend on the initial frame-wise prediction for boundary localization, which require high-quality classification as a fundamental step.

\footnotetext[1]{\href{https://github.com/adjkamel/segsim}{\url{https://github.com/adjkamel/segsim}}}

To tackle the previous limitations, in this paper, we propose ASESM (Action Segmentation via Explicit Similarity Measurement), a supervised learning architecture to enhance action segmentation by measuring the similarity between frames explicitly during the training and testing process.  The challenge of inaccurate initial prediction is addressed by using frame-level multi-resolution features as input to multiple transformer encoders to capture global details in small scale features and local details in large scales. Unlike \citet{singhania2021coarse} that use multi-resolution features of the input then stack them together for further processing, we process each of the scaled features with an independent encoder to diverse the feature extraction and predictions. The resulting predictions undergo frame-wise similarity voting to identify the most likely correct class. To address the lack of explicit boundary localization, we introduce a new iterative boundary correction algorithm that adjusts boundary locations. Unlike existing methods that rely solely on initial frame-wise prediction for boundary correction, our approach involves the input frame-wise features around the boundary area to measure the similarity between consecutive frames in an iterative process. The boundary correction is applied during each training iteration,  adapting the learning process to adjust the boundaries. The corrected prediction is further refined through multiple stages using temporal convolutions. At prediction time, as a post-processing phase, we optionally apply boundary correction again, followed by a segment smoothing technique that removes outlier classes within segments by measuring the similarity between consecutive predictions, to mitigate the impact of wrong frame-wise classification.

Beyond supervised action segmentation, we propose a fully unsupervised boundary detection-correction algorithm that leverages the same similarity measurement metrics used in our supervised approach. This unsupervised technique identifies segment boundaries directly from frame-wise features without requiring any training or coarse initial boundaries. Our experimental results show that both supervised and unsupervised algorithms achieve superior or on par accuracy compared to existing approaches. Moreover, the unsupervised algorithm outperforms other techniques, including those utilizing representation learning. One of the main advantages of our supervised similarity measurement method is that it can be integrated into various existing backbones to improve their segmentation accuracy. The ablation studies demonstrate the significant impact of the proposed similarity measurement components on overall performance. The contributions of our work can be summarized as follows: 

\begin{figure*}
\centering
\includegraphics[width=0.7\linewidth]{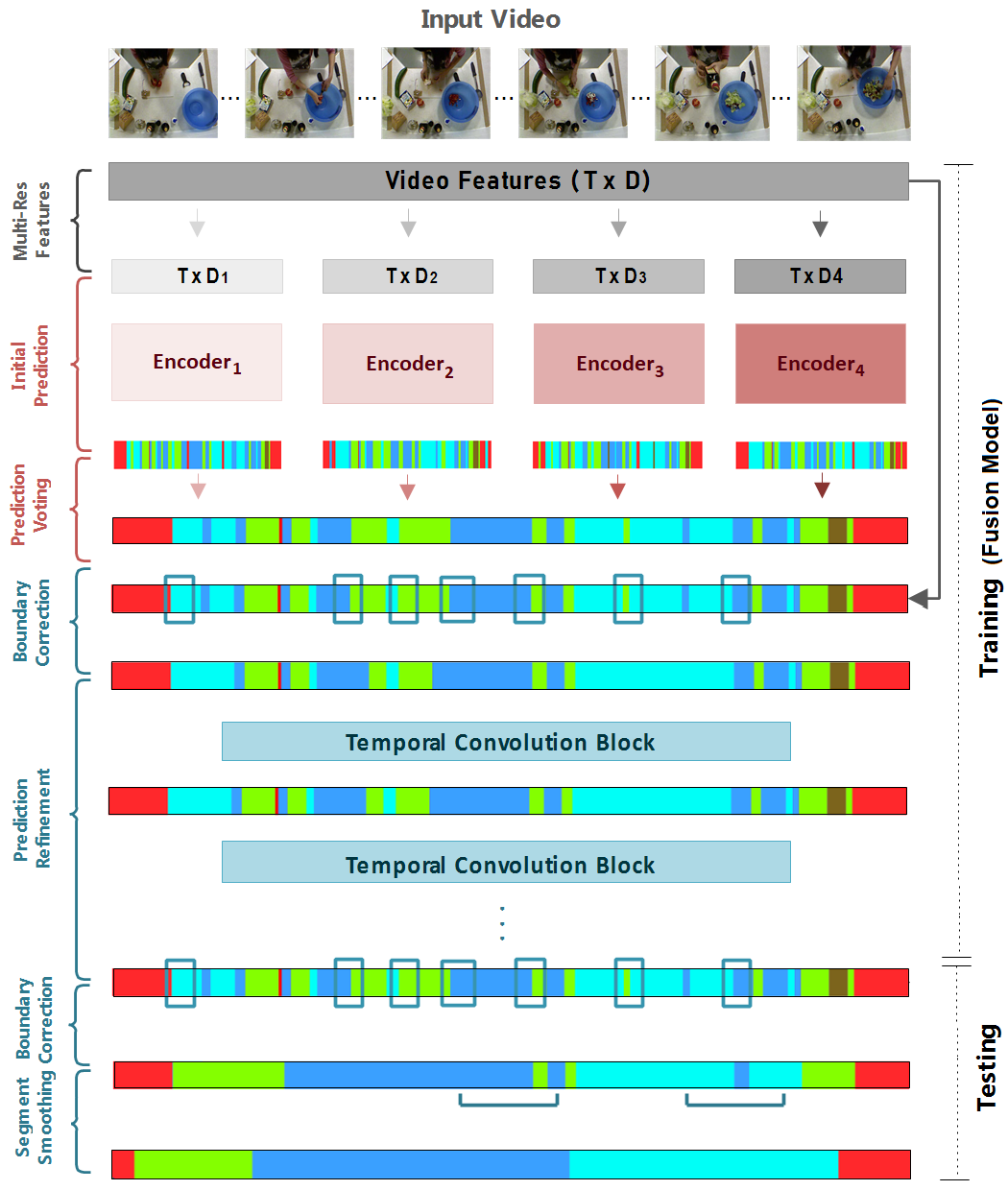}
\caption{Framework of our supervised learning architecture for explicit similarity measurement. We involve three levels of similarities: Multi-resolution prediction similarity voting, boundary correction based on frame-wise feature similarity, and segment smoothing based on frame-wise prediction similarity. }
\label{fig:frame} 
\end{figure*}

\begin{enumerate}[label={(\arabic*)}]

\item Enhancing action segmentation by explicitly measuring the similarity between features and predictions in a supervised learning framework through i) voting across multiple frame-wise predictions generated form features of different resolutions, ii) boundary correction based on feature similarity, and ii) segment smoothing to remove outlier classes within segments.

\item A new boundary correction algorithm that iteratively adjusts boundaries during training using initial predictions, incorporated with frame-wise feature similarity measurement to explicitly improves the accuracy of boundary localization.

\item  Based on the same similarity metrics of the boundary correction algorithm, a new fully unsupervised boundary detection-correction algorithm is proposed. This algorithm directly leverages the similarity between features of consecutive frames to identify  and refine segment boundaries without requiring a learning scheme.

\end{enumerate}

The rest of this paper is organized as follows: Section \ref{section2} reviews related works. Section \ref{section3} introduces the technical details of our method. Section \ref{section4} presents the analysis of the experimental results and ablation study, followed by a conclusion in Section \ref{section5}.

\section{Related work}  \label{section2}

\textbf{Feature extraction:} The first methods for feature extraction are based on hand-craft features, where dense trajectories \cite{wang2011aser, wang2013action} is a commonly used approach. It involves tracking key points across consecutive frames using optical flow with space-time descriptors such as Histograms Of Gradients (HOG), Histograms of Optical Flow (HOF), and Motion Boundary Histogram (MBH). After the introduction of deep learning, 3D CNN has been widely used for feature representation, either for individual frames or a sequence of frames. Specifically, state-of-the-art Inflated 3D CNN (I3D)  \cite{carreira2017quo} employs the Inception-V1 model as a backbone \cite{szegedy2015going} with 3D kernels to facilitate the direct spatio-temporal processing. The I3D model is pre-trained on large-scale video datasets that contain a diverse range of human actions \cite{kay2017kinetics}. The pre-training enables the model to learn generic features useful for various downstream tasks, including action recognition and video segmentation.

\textbf{Supervised action segmentation:} 
Supervised action segmentation methods mainly utilize frame-wise labeled videos for supervised learning, where Temporal 1D Convolutional Networks (TCNs) have gained widespread adoption due to their ability to effectively capture temporal dependencies within sequences. Diverse architectures were introduced a temporal convolution encoder-decoder architecture that decreases and increases the size of the temporal resolution using pooling and up-sampling, respectively \cite{lea2017temporal, ding2018weakly, lei2018temporal,singhania2023c2f}. However, the multi-stage architecture (MS-TCN) retains the same temporal resolution and extends the receptive field by employing increasingly larger dilated convolutions. This approach avoids using pooling to preserve features, as boundaries are sensitive to feature loss. 

Transformer which was originally designed for natural language processing, mainly relies on attention mechanism for sequence modeling \cite{vaswani2017attention}. In action segmentation, TimeSformer \cite{bertasius2021space} adapts the traditional Transformer architecture for video processing \cite{dosovitskiy2020vit, arnab2021vivit} to learn spatio-temporal features directly from sequences of frame-level patches. Their experiments indicate that using separate spatial and temporal attention is more efficient.  ASFormer \cite{yi2021asformer}  was among the pioneering transformer architectures for temporal action segmentation. It adapts the encoder-decoder framework of ED-TCN \cite{lea2017temporal}, and replaces the convolutions operations with transformer blocks with local window attention that grows in size with each layer. \blue{TUT \cite{du2023we} proposed a pure Transformer-based model that incorporates temporal sampling instead of temporal convolutions to reduce complexity, and to address the boundary misclassification, by proposing a boundary-aware loss that leverages similarity scores from attention modules.}

Although action segmentation is widely applied on RGB video sequences, some works investigated the problem in skeleton-based sequences \cite{filtjens2022skeleton,tian2023stga}. Graph Convolutional Network (GCN) is more appropriate to model the skeleton. Therefore, they are widely utilized in the segmentation process instead of temporal convolutions.  Multi-Stage spatial-temporal Graph Convolutional Neural network (MS-GCN) \cite{filtjens2022skeleton} replaces initial temporal convolutions with spatial graph convolutions to capture spatial hierarchies and long-term temporal dynamics in the skeleton sequence. STGA-Net \cite{tian2023stga} introduces a spatial-temporal graph attention network (STGA-Net), which includes an attentive block within the encoder-decoder to model dynamic correlations among human joints, addressing the lack of an explicit transition rule between segments.

\textbf{Over-segmentation:} The prediction of a single action class per frame frequently encounter over-segmentation errors, where boundary segments are misallocated. Therefore, alternative methods specifically tackle the problem by focusing on boundary detection or refinement. BCN \cite{wang2020boundary} mitigates boundary ambiguity by introducing a refinement branch for MS-TCN to detect the start and the end of segments. \citet{ishikawa2021alleviating} proposed integrating an additional network branch dedicated to boundary identification by assigning high probability to boundary frames. \citet{ding2022leveraging} and \citet{ding2018weakly} proposed using soft boundary detection. Instead of having a clear-cut distinction of where one action ends and another begins, they detect fluid transitions for more robust boundary localization.

\textbf{Unsupervised action segmentation:} Unsupervised action segmentation approaches can be categorized into two categories. Self-supervised methods that learn action representations, and fully unsupervised methods that operates on the input features without requiring a learning framework. In self-supervised approaches, \citet{kukleva2019unsupervised} exploit the sequential nature of activities to learn continuous temporal embedding based on frame-wise features with respect to their relative time in the sequence, and then the embedding features are clustered to identify temporal segments. \citet{vidalmata2021joint} combine visual embedding derived from a predictive U-Net architecture with a temporal continuous embedding. Object-centric Temporal Action Segmentation (OTAS) \cite{li2024otas} introduced self-supervised global and local feature extraction with a boundary selection module to detect salient boundaries. In contrast to existing approaches that involve representation learning then offline clustering, \citet{kumar2022unsupervised} proposed a joint self-supervised representation learning and online clustering in a unified end-to-end learning scheme. Although fully unsupervised methods are not widely investigated as self-supervised methods, there are some works  showed that applying similarity metrics or clustering directly on the frame-wise features could outperform representation learning-based methods \cite{sarfraz2019efficient,sarfraz2021temporally,du2022fast}. 

\textbf{Weakly supervised action segmentation:} Weakly supervised approaches involve global labeling such as action sets, transcript, or timestamp methods. In action sets methods,  each video is associated with a unique unordered set of actions.  Transcript methods require an ordered lists of actions, and  timestamp methods rely on pseudo frame-wise labels. \citet{richard2018action} proposed one of the first action sets methods that doesn't require prior knowledge of the number or the occurrence order of actions. \citet{lu2022set} exploit the fact that videos within the same task have similar action orderings, to introduce the Pairwise Ordering Consistency (POC) loss that ensures consistent predictions across different videos of the same task. NN-Viterbi \cite{richard2018neuralnetwork}, a pioneering transcript-based method, which demonstrates higher performance by generating pseudo-labels from transcripts using Viterbi decoding for model training. \citet{xu2024efficient}  filter noisy boundaries and detect transitions while incorporating video-level losses to improve semantic learning for noisy pseudo-segmentations. \citet{du2023timestamp} introduce a clustering-based framework for timestamp-supervised action segmentation, which tackles incorrect pseudo-labels in ambiguous intervals. This framework includes pseudo-label ensembling to generate high-quality labels and iterative clustering for their propagation. \citet{hirsch2024random} reformulate temporal action segmentation as a graph segmentation problem with weak supervision via timestamp labels to reduce annotation costs.

\begin{figure}
\centering
\includegraphics[scale=0.6]{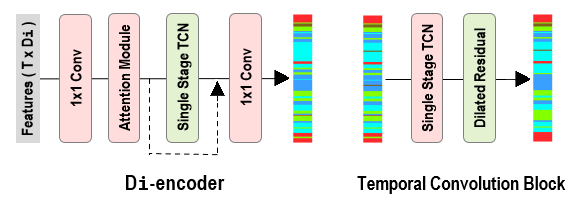}
\caption{Encoder structure (left) and Temporal Convolution Block:TCB (right), which are modified versions of the ASFormer \cite{yi2021asformer} encoder and a single stage of MS-TCN \cite{farha2019ms}, respectively. The components added are in green color. }
\label{fig:comp} 
\end{figure}

\section{Methodology}\label{section3}
This section describes the framework of the proposed supervised action segmentation, which requires initial predictions to identify initial boundaries. The framework includes feature extraction, similarity voting, boundary correction, prediction refinement, and segment smoothing. At the end of this section,  the proposed fully unsupervised detection-correction method is presented, as an independent algorithm that requires only the raw frame-wise features.

\subsection{Multi-resolution features}
The framework of our proposed supervised action segmentation method is presented in Fig. \ref{fig:frame}. The first step is frame-wise feature extraction. Following the previous works \cite{farha2019ms,yi2021asformer,farha2019ms},  we use the features generated by I3D \cite{carreira2017quo} model. 

Given a video $V=\{x_1,x_2,...,x_T\}$ of a sequence of frames, where $T$ is the number of frames, and $x_i$ is a single frame. Let $f=\{f_1,f_2,f_3,...,f_T\}$ the corresponding features sequence generated by the I3D model, where $f_i \in \mathbb{R}^D$ are the features of each frame, with  $D$ the feature dimension. Existing methods down-sample the feature $D$ to a specific frame-wise feature dimension such as 64 then process the sequence. However, multiple global and local informative features can be found in different resolutions. Therefore, we down-sample the features $f$ to four different resolutions $R_i \in \mathbb{R}^{T\times D_i}$,  $i=\{1,2,3,4\}$ and $D_i=\{32,64,128,256\}$, using 1D convolution with a $1\times 1$ kernel and a stride of 1 to maintain the same number of frames as the input.

\begin{equation}
R_i = Conv1D(f,D_i,str=1,ker=1)
\end{equation}

\begin{figure}
\centering
\includegraphics[scale=0.45]{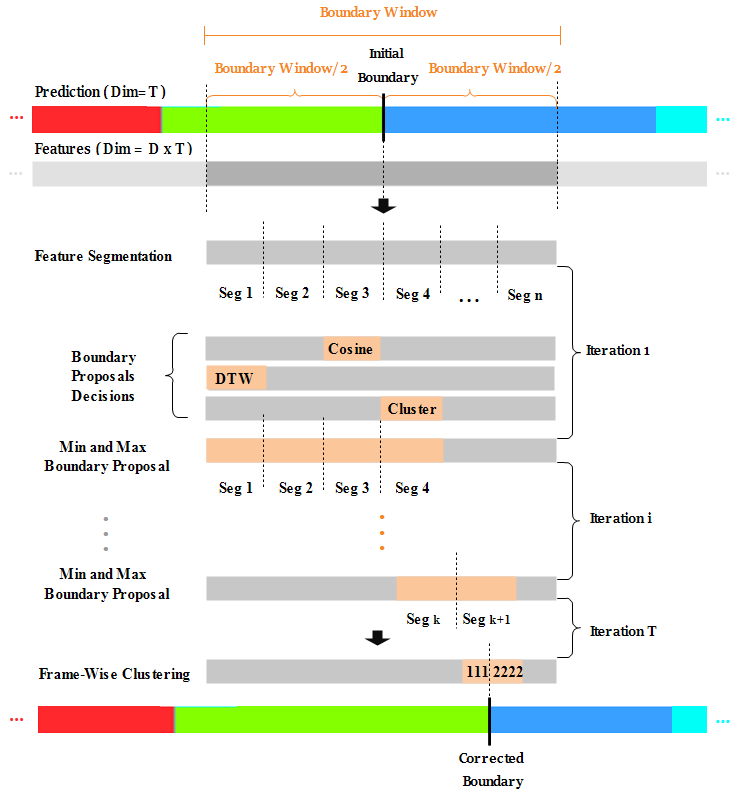}
\caption{Visual illustration of the boundary correction algorithm.}
\label{fig:bound} 
\end{figure}

\subsection{Feature extraction and initial prediction}
Each of the four feature sequence resolutions $R_1$,$R_2$,$R_3$, and $R_{4}$ is processed with a separate encoder for temporal modeling and frame-wise label prediction. We denote $P_i$ as the frame-wise prediction from each encoder:
\begin{equation}
P_i=Encoder_{i}(R_i)
\end{equation}

We adopt the encoder structure of ASFormer \cite{yi2021asformer}, which consists of a self-attention layer followed by feed-forward layer. Unlike their original encoder structure, the point-wise fully connected layer in the feed-forward layer is replaced with a dilated temporal convolution. We further improve the encoder structure by incorporating a single-stage temporal convolution block of MS-TCN \cite{farha2019ms} before the feed-forward layer. This block consists of two 1D convolution layers with four residual dilated convolutions in between. The new modified encoder is shown in Fig. \ref{fig:comp}(left). The added single-stage temporal convolution is surrounded by a residual connection in $Encoder_{2}$,$Encoder_{3}$, and $Encoder_{4}$. However, we don't use residual connection around the single-stage temporal convolution in $Encoder_{1}$.

\begin{algorithm}[t]
\footnotesize	
\caption{\small Supervised Boundary Correction}\label{algo:bndcorr} 
\begin{algorithmic}[1]
\vspace{1.2mm}
    \State \textbf{Input:} $P_{init}$: Initial Predictions of size = $T$\\\vspace{1.2mm}
           \quad\quad\quad$V_{feat}$: Video Features of size = $T \times D$\vspace{1.2mm}
    \State\textbf{Output:} $P_{corr}$: Corrected Predictions \vspace{1.2mm}
    \State$B_{win}$= $ MaxDst(Bound(P_{init}))/MinDst(Bound(P_{init}))$ \vspace{1.2mm}
    \State$B_{seg}=B_{win}/min(div(B_{win}))$\vspace{1.2mm}
    \For{$i$ \textbf{in} $Bound(P_{init})$} \vspace{1.2mm}
    \State $B_{feat}= V_{feat}[i-B_{win}/2  ,  i+B_{win}/2]$\vspace{1.2mm}
    \Repeat\vspace{1.2mm}
    \State $ Seg_{feat}=Segment(B_{feat}, B_{seg})$\vspace{1.2mm}
    \State $B_{Cosine}=Min(Cosine(Seg_{feat}))$\vspace{1.2mm}
    \State $B_{Dtw}=Max(Dtw(Seg_{feat}))$\vspace{1.2mm}
    \State $B_{Clust}=Cluster(Seg_{feat},2)$\vspace{1.2mm}
    \State $B_{Start}=Min(B_{Cosine},B_{Dtw},B_{Clust})$\vspace{1.2mm}
    \State $B_{End}=Max(B_{Cosine},B_{Dtw},B_{Clust})$\vspace{1.2mm}
    \State $B_{feat}=V_{feat}[B_{Start},B_{End}]$\vspace{1.2mm}
    \Until{$||B_{feat}||=B_{seg}$ }\vspace{1.2mm}
   \State $B_{idx}=Cluster(B_{feat},2)$\vspace{1.2mm}
   \State $P_{corr}=Correct(P_{init},B_{idx})$\vspace{1.2mm}
    \EndFor
\end{algorithmic}
\label{alg:corr}
\end{algorithm}
\normalsize

\subsection{Prediction similarity voting}

Existing methods \cite{farha2019ms,yi2021asformer} show that employing multi-stage refinement of the initial prediction plays a vital role in the frame-wise prediction and segmentation process. However,  the refinement process mainly rely on  the initial prediction, and some of the predictions of misclassified frames remains the same along the rest of the refinement process. Therefore, we select the most likely correct class from the four encoders' predictions using frame-wise majority voting to obtain accurate  initial prediction based on multiple resolutions. If two or more encoders assign the same class to a specific frame, that class is considered correct. If all the encoders assign different classes, the prediction from $Encoder_{4}$ is trusted as the correct class due to the diversity of features in its input sequence. We denote $P_{init}$ as the initial prediction obtained through voting, which is used for the rest of the refinement process: 
\begin{equation}
P_{init} = Voting(P_1, \dots, P_4)
\end{equation}

\subsection{Boundary correction}
Using encoders with an attention mechanism for initial prediction and temporal convolutions for prediction refinement greatly enhances the segmentation through frame-wise classification. However, this doesn't explicitly guarantee high accuracy around boundary areas. Therefore, we propose incorporating a boundary correction algorithm during training before applying prediction refinement on $P_{init}$.

As illustrated in Fig. \ref{fig:bound} and detailed in Algorithm \ref{algo:bndcorr}, the boundary correction process starts by identifying boundaries in the initial prediction $P_{init}$. For each detected boundary, the corresponding frame index is identified in the input feature sequence, then a boundary window $B_{win}$ of features around the boundary frame is selected and segmented  into smaller, equal-sized boundary segments $B_{seg}$. Three similarity measurement techniques are applied to these boundary segments; binary clustering, cosine similarity, and binary clustering. We denote $seg_{feat}=\{s_1,s_2,\dots,s_m\}$, the feature sequence of the boudany window segments. The Cosine similarity and Dynamic Time Warping (DTW) similarity are calculated between each two consecutive segments. The minimum Cosine value indicates low similarity, and the maximum DTW value indicates low similarity. In both cases, the corresponding segment is considered to contain an action transition:

\small
\begin{align}
\begin{split}
SimCosine&=\{Cosine(s_{0},s_{1}), \dots Cosine(s_{m-1},s_{m})\}\\
B_{Cosine}&=argmin(SimCosine) \\
\label{eq1}
\end{split}
\end{align}
\begin{align}
\begin{split}
SimDtw&=\{Dtw(s_{0},s_{1}), \dots Dtw(s_{m-1},s_{m})\}\\
B_{Dtw}&=argmax(SimDtw)
\label{eq1}
\end{split}
\end{align}
\normalsize
Where $Bcos_i$ and $Bdtw_i$ are the indexes of the boundary segment suggested by the low similarity of Cosine and DTW, respectively. $Cosine$ and $DTW$ are calculated between the feature sequence of frames in $seg_{i-1}$ and the features sequence of frames in $seg_{i}$. For binary clustering (0 or 1), the transition can be detected by finding the corresponding longest clustered subsequence  $C_l$ among all clustered subsequences $\{C_1,\dots, C_k, \dots, C_K\}$  that contain a sequence of 0 followed by a sequence of 1: 

\begin{align}
\begin{split}  
C_k=\{c_{1},\dots,c_{j-1},c_{j},\dots, c_{n}\} \\ 
c_j =
\begin{cases} 
0, &  j \in [1, j-1] \\
1, &  j \in [j, n]
\end{cases}\\
C_{L} = \max(\|C_1\|, \|C_2\|, \dots, \|C_K\|)\\
B_{Clust}= i \quad \text{where} \quad   C_{L}(c_i) \neq C_{L}(c_{i-1})
\end{split}
\end{align}
Where $B_{Clust}$ is the index of the corresponding transition segmentation. The indexes obtained from $B_{Cosine}$, $B_{Dtw}$, and $B_{Clust}$, are used as proposals of segments where boundaries are likely to exist. The three proposal segments and their intervening segments are concatenated to form a new reduced feature sequence input. This process is iteratively repeated, with each iteration the boundary window is segmented into smaller segments until the size of the boundary window matches the size of the boundary segment. Finally, binary clustering is applied on the frames of the last boundary segment to determine the final boundary. The resulting corrected predictions $P_{corr}$ can be formulated as :
\begin{equation}
P_{corr}=BCorr(P_{init},B_{Cosine}, B_{Dtw}, B_{Clust} )
\end{equation}
Where $BCorr$ is the boundary correction algorithm. $P_{corr}$ is further refined by multiple iterations of Temporal Convolution Blocks (TCB). Unlike the model structure proposed by MS-TCN \cite{farha2019ms}, each of our TCB blocks consists of a single-stage temporal convolution followed by dilated residual layers (Fig. \ref{fig:comp}(right)). We use 4 TCBs, where the output prediction of each block is served as input for the next block. The resulting refined prediction can be written as: 
\begin{equation}
P_{refin}=TCB(...(TCB(P_{corr})))
\end{equation}


\begin{figure}
\centering
\includegraphics[width=1\linewidth]{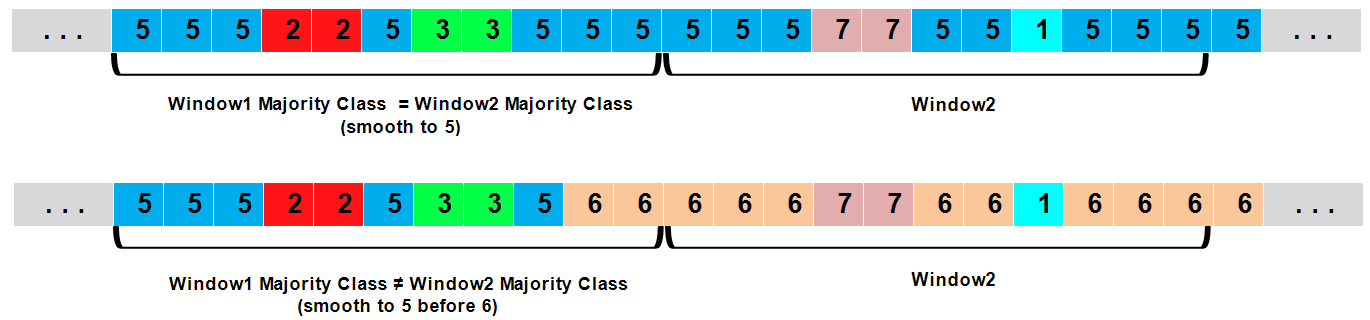}
\caption{Segment smoothing: Two shifting windows are used. The smoothing happens within the first window only. The second window is used to check the location of the boundary.}
\label{fig:smooth} 
\end{figure}

\subsection{Segment smoothing}
Removing outlier classes within segments of $P_{refin}$ improves the accuracy, we propose to use two consecutive windows that are shifted from left to write together. The smoothing happens in the first window. The second window is only used to check whether the next frames are included in the current segment or a new segment starts. In other words, it is used to check the boundary location. As shown in Fig. \ref{fig:smooth}, there are two possible cases while shifting the two windows. In the first case, both windows have the same majority prediction, which results in replacing all the classes in the segment with the majority class. In the second case, the majority class in the first window is different from the majority class in the second window, which means that the second window falls into a new segment, and the classes of the first window are replaced with the majority class except for the frames that have the same class as the majority class label of the second window. The final frame-wise prediction of the framework is obtained after smoothing:
\begin{equation}
P_{final}=Smth(P_{refin}) 
\end{equation}





\subsection{Model training and testing}
Each of the four encoders $Encoder_i$ is trained individually. The Fusion model includes loading the pre-trained four encoders, generating frame-wise prediction from each of them, applying the voting, boundary correction, and refining the prediction through multiple iterations using TCB. The voting and boundary correction are applied alongside the training process in each learning iteration. In the testing phase, as a post-processing step,  boundary correction is applied again (optionally), followed by segment smoothing.

\normalsize
\subsection{Unsupervised boundary detection-correction}
Building on the same idea of using clustering, Cosine, and DTW for similarity measurement, we propose an unsupervised boundary detection-correction algorithm that operates without any initial coarse prediction. The goal of this algorithm is to detect initial boundaries and refine them based on the similarity between consecutive frames features. Given a sequence of video features $f=\{f_1,f_2,f_3,...,f_T\}$, we first apply 1D convolution to reduce the frame-wise feature size and obtain $V_{feat}$, as our input of the algorithm. Next, we generate a list of possible boundaries within the sequence using clustering, Cosine, and DTW between consecutive frame features. The clustering is performed across all frames based on the number of classes.

The application of the three similarity techniques generally results in a higher number of boundary proposals than the real boundaries. Therefore, for the boundaries proposed by Cosine or DTW, consecutive boundaries with similarity values below or above the mean similarity value are removed, respectively. Then, nearby boundaries, based on a specified boundary interval $B_{intrv}$, are also removed. For boundaries proposed by clustering, only  close boundaries according to the given interval $B_{intrv}$ are removed. Finally, the refined boundaries from all three similarity methods are merged, and the final boundaries are determined by averaging those that fall within the threshold of the specified boundary interval $B_{intrv}$. Algorithm \ref{algo:bndunsp} formalizes the previous explanation.
\begin{algorithm}
\caption{\small Unsupervised Boundary Detection-Correction}\label{algo:bndunsp}
\footnotesize	
\begin{algorithmic}[1]
\vspace{1.2mm}
    \State \textbf{Input:} $V_{feat}$: Video Features of size = $T \times D$\\ \vspace{1.2mm}
\quad\quad\quad$B_{intrv}$: Threshold Interval Between Boundaries (frames)\vspace{1.2mm}    
     \State\textbf{Functions:} $RmvBounds$: Remove boundaries if gap < $B_{intrv}$\\ \vspace{1.2mm}
     \quad\quad\quad\quad\quad$SubMean$: Keep boundaries < mean of all distances \\ \vspace{1.2mm}  
     \quad\quad\quad\quad\quad$SupMean$: Keep boundaries > mean of all distances \\ \vspace{1.2mm}  
     \quad\quad\quad\quad\quad$Merge$: Consider all boundaries \\ \vspace{1.2mm}  
     \quad\quad\quad\quad\quad$MeanBounds$: Consider the mean if gap < $B_{intrv}$  \vspace{1.2mm}  
    \State\textbf{Output:} $B_{list}$: List of boundaries \vspace{1.2mm} 
    \State \quad$B_{Clust}=Cluster(V_{feat},num\_classes)$\vspace{1.2mm}
    \State \quad$B_{Clust}=RmvBounds(B_{Clust}, B_{intrv})$\vspace{1.2mm}
     \State\quad $B_{Cosine}=SubMean(Cosine(V_{feat}))$\vspace{1.2mm}
     \State \quad$B_{Cosine}=RmvBounds(B_{Cosine}, B_{intrv})$\vspace{1.2mm}
    \State \quad$B_{Dtw}=SupMean(Dtw(V_{feat}))$\vspace{1.2mm}
    \State \quad$B_{Dtw}=RmvBounds(B_{Dtw}, B_{intrv})$\vspace{1.2mm}
    \State \quad$B_{merge}=Merge(B_{Cosine},B_{Dtw},B_{Clust})$\vspace{1.2mm}
    \State \quad$B_{list}=MeanBounds(B_{merge}, B_{intrv})$\vspace{1.2mm}
\end{algorithmic}
\label{alg:detcorr}
\end{algorithm}

\section{Experiments}\label{section4}

\subsection{Datasets and evaluation}
We evaluate our method on 50Salads, GTEA, and Breakfast datasets. 50Salads dataset consists of 50 videos of 17 actions focused on salad preparation tasks in the kitchen, performed by 25 actors. Each actor prepares two distinct salads. The videos are recorded from a top-down view, average 20 actions, and lasts for 6 minutes and 4 seconds. GTEA dataset comprises 28 videos of 7  activities with 11 action classes such as coffee or cheese sandwich preparation by 4 actors. The videos were recorded using a camera mounted on the actor's head. Each video contains 20 instances of actions. Breakfast dataset contains 1712 videos of various breakfast preparation activities in 18 different kitchen settings with a total of 48 action classes. Each video includes an average of 6 actions.

Following the previous methods,  we employ five-fold cross-validation for 50Salads dataset and four-fold cross-validation for GTEA and Breakfast datasets. We assess the performance using the three key metrics: frame-wise accuracy (Acc), edit distance (Edit), and segmental F1 score at overlapping thresholds of 10\%, 25\%, and 50\%, denoted as F1@10, F1@25, and F1@50, respectively. In the case of the unsupervised detection-correction algorithm, we assess the performance using the F1@10 score.

\subsection{Implementation details}

In Algorithm \ref{algo:bndcorr}, we set $B_{win}=16$ frames and $B_{seg}=4$ frames for the 50Salads and Breakfast datasets, and $B_{win}=8$ and $B_{seg}=4$ for the GTEA dataset. Moreover, we found that changing the boundary window has no noticeable impact on the performance. In the ablation study, we investigate manually specified smoothing window sizes as well as an automatic smoothing window size $S_{win}=MaxDistBound/10$, where $MaxDistBound$ is the maximum distance (frames) between two consecutive boundaries in the video. We set the boundary interval threshold manually based on examining the segmentation results on one sample. We also propose to use an automatically calculated threshold as $B_{intrv}=MaxDistBound(Cosine,Clust,DTW)$, which refers to the maximum distance between boundaries decided by Cosine, Clustering, and DTW similarity measurements techniques. Each of the four encoders and the fusion model are trained for 50 epochs with a batch size of 1 video in each learning iteration, with a learning rate of 0.0005, and Adam optimizer. The experiments were conducted using the PyTorch framework on a machine with an Nvidia GeForce RTX 4090 GPU, 64GB of RAM, and an Intel(R) Core(TM) i9-13900k CPU.

\begin{table*}
\footnotesize
\setlength{\tabcolsep}{1.2pt}
\renewcommand{\arraystretch}{1.3}
\caption{Comparison of supervised action segmentation results with the state-of-the-art on 50Salads and GTEA datasets with S\textsubscript{win}=80 and S\textsubscript{win}=4, respectively. Results obtained with boundary correction during training. InTestCorr: Apply boundary correction of initial prediction at test time. PostTestCorr: Apply boundary correction in post-processing.}
\centering
\begin{tabular}{l|lllll|lllll|lllll}
\hline
&&&50Salads&&&&&GTEA&&&&& \blue{Breakfast}&\\
Method   &F1@10&F1@25&F1@50&Edit&Acc&F1@10&F1@25&F1@50&Edit&Acc&\blue{F1@10}&\blue{F1@25}&\blue{F1@50}&\blue{Edit}&\blue{Acc}\\    
\hline 

MS-TCN \cite{farha2019ms} (CVPR'19) & 76.3 & 74.0 & 64.5 & 67.9 & 80.7 &  85.8 & 83.4 & 69.8 & 79.0 & 76.3&  \blue{52.6} & \blue{48.1} & \blue{37.9} & \blue{61.7} & \blue{66.3} \\
MS-TCN++ \cite{li2020ms} (CVPR'20) & 80.7 & 78.5 & 70.1 & 74.3 & 83.7 &88.8 & 85.7 & 76.0 & 83.5 & 80.1&  \blue{64.1}& \blue{58.6}& \blue{45.9} &\blue{65.6} &\blue{67.6} \\
SSA-GAN \cite{gammulle2020fine} (PR'20) & 74.9 & 71.7 & 67.0 & 69.8 & 73.3 & 80.6 & 79.1 & 74.2 & 76.0 & 74.4& \blue{-}  &  \blue{-}  &  \blue{-}  &  \blue{-} & \blue{-}  \\
BCN \cite{wang2020boundary} (ECCV'20) & 82.3 & 81.3 & 74.0 &74.3& 84.4 &88.5 & 87.1 &  77.3 &84.4 &79.8&  \blue{68.7}& \blue{65.5}& \blue{55.0} &\blue{66.2}& \blue{70.4}\\
ASFormer \cite{yi2021asformer} (BMVC'21) & 85.1 & 85.4 & 79.3 &81.9 &85.9&90.1 & 88.8 & 79.2 & 84.6 &79.7&  \blue{76.0} &\blue{70.6}& \blue{57.4}& \blue{75.0}& \blue{73.5}\\
DTGRM \cite{wang2021temporal} (AAAI'21) & 79.1 & 75.9 & 66.1 &72.0 &80.0 &  87.8 & 86.6 & 72.9 &83.0 &77.6 & \blue{68.7} & \blue{61.9} & \blue{46.6} & \blue{68.9} & \blue{68.3} \\
ASRF \cite{ishikawa2021alleviating} (WACV'21) & 84.9 & 83.5 & 77.3 & 79.3 & 84.5& 89.4 & 87.8 & \underline{79.8} & 83.7 & 77.3& \blue{74.3} &  \blue{68.9} &  \blue{56.1} & \blue{72.4} & \blue{67.6} \\

UVAST \cite{behrmann2022unified} (ECCV'22) & 86.2& 81.2& 70.4& 83.9 &79.5  &   77.1 &69.7 &54.2& 90.5& 62.2  &  \blue{\underline{76.7}} &\blue{70.0}& \blue{56.6}& \blue{77.2} &\blue{68.2}    \\
LTContext\cite{bahrami2023much} (ICCV'23) & 89.4 & 87.7 & \underline{82.0} &83.2 &86.0&  - & - & - & - & - & \blue{77.6} & \blue{\textbf{72.6}} & \blue{\textbf{60.1}} & \blue{\textbf{77.0}} & \blue{74.2}\\
\blue{LGTNN \cite{tian2023local} (MMS'23)}  &  \blue{87.5} & \blue{86.2} & \blue{79.8} & \blue{82.0} & \blue{86.1}  & \blue{91.5}& \blue{90.4} & \blue{80.3} & \blue{87.5} &\blue{\textbf{81.2}}&  \blue{76.2}& \blue{\underline{71.5}} &\blue{57.5} &\blue{75.2} &\blue{72.5} \\
\hline
\textbf{ASESM} (InTestCorr)  & \underline{89.3} & \underline{88.1} & 81.6 & \underline{83.8} &\underline{87.4}& \textbf{91.7}& \textbf{90.6}  & \textbf{82.0} & \textbf{88.0}  & \underline{80.5}&  74.7   &      69.7    &    57.6 & 72.5  & 75.5             \\
\textbf{ASESM} (InTestCorr+PostTestCorr)  & \textbf{89.5} & \textbf{88.3} & \textbf{82.5} & \textbf{83.9} &\textbf{88.0}&\underline{91.7} & \underline{90.4} & 78.9 & \underline{87.9} &78.9&  75.3 & 70.1  &  \underline{58.0} & 73.6  & \textbf{75.7}\\
\hline
\end{tabular}
\label{table:sup}
\end{table*}

\begin{table}
\small
\renewcommand{\arraystretch}{1.3}
\setlength{\tabcolsep}{6pt}
\caption{Comparison of unsupervised action segmentation results with the state-of-the-art methods on Breakfast dataset.}
\centering
\begin{tabular}{l l l }

\hline

    &\multicolumn{2}{c}{F1}  \\

Method   &Breakfast & \blue{50salads}  \\

\hline

CTE \cite{kukleva2019unsupervised} (CVPR'19) &26.4& - \\
DGE \cite{dimiccoli2020learning} (TIP'20) &51.7 & - \\
UDE \cite{swetha2021unsupervised} (ICIP'21) &31.9 & \blue{34.4}\\
TW-FINCH \cite{sarfraz2021temporally} (CVPR'21) &49.8 & -\\
TOT \cite{kumar2022unsupervised} (CVPR'22)&31.0& \blue{48.2} \\
ABD \cite{du2022fast} (CVPR'22) &52.3& - \\
TSA \cite{bueno2023leveraging} (CVPR'23)  &58.0& - \\
\blue{SaM \cite{xing2024unsupervised} (AAAI'24)}  & \blue{55.9} & \blue{-} \\

\textbf{ASESM} (B\textsubscript{intrv}=300) & \textbf{63.2} & \blue{\textbf{51.2}}  \\

\hline
\end{tabular}
\label{table:unsup}
\end{table}

\begin{table}
\footnotesize
\caption{Comparison of unsupervised action segmentation results on F1 score with a state-of-the-art semi-supervised method on 50Salads and GTEA datasets.}
\scriptsize
\centering
\setlength{\tabcolsep}{3.3pt}
\renewcommand{\arraystretch}{1.7}
\begin{tabular}{l l l  l}
\hline
Method   &Labels\% & 50Salads  & GTEA \\
\hline 
SemiTAS \cite{ding2022leveraging} (ECCV'22)  &   5 & 37.4&59.8\\
SemiTAS \cite{ding2022leveraging} (ECCV'22)  &  10 & 47.3&  \textbf{71.5} \\
\textbf{ASESM} (B\textsubscript{intrv}=auto)&    \textbf{0} & 43.7 & 57.0 \\
\textbf{ASESM} &   \textbf{0} & \textbf{51.2} (B\textsubscript{intrv}=500) & \underline{70.4} (B\textsubscript{intrv}=70) \\
\hline
\end{tabular}
\label{table:semi}
\end{table}

\begin{table}
\footnotesize
\caption{Comparison of unsupervised and supervised action segmentation results on F1@10 score with timestamp  methods on 50Salads, GTEA, and Breakfast datasets.}
\scriptsize
\centering
\setlength{\tabcolsep}{3.3pt}
\renewcommand{\arraystretch}{1.7}
\begin{tabular}{l  l  l l}
\hline
Method   & 50Salads  & GTEA & Breakfast\\
\hline 
\blue{RWS \cite{hirsch2024random} (WACV'24)} & \blue{76.7}  & \blue{80.9} & \blue{70.9}\\
\blue{UVAST \cite{behrmann2022unified} (ECCV'22) (timestamp)} & \blue{75.7} & \blue{70.8} & \blue{72.0} \\
\textbf{ASESM} (unsupervised)  & 51.2  & 70.4  & 63.2\\
\textbf{ASESM} (superivsed)  & 89.5  & 91.7  &75.1\\
\hline
\end{tabular}
\label{table:timestamp}
\end{table}

\begin{table*}
\small
\caption{Ablation study on the effect of each component of the supervised framework on the performance. Results obtained on split 1 of 50Salads dataset.}
\centering
\setlength{\tabcolsep}{0.6mm}
\renewcommand{\arraystretch}{1.3}
\begin{tabular}{c  c c c c c c c c c cc c c c c}
\hline
 Enc\textsubscript{1} &Enc\textsubscript{2} &Enc\textsubscript{3} &Enc\textsubscript{4} &TCB &Vot & TrainCorr& InTestCorr &PostTestCorr & Smooth &F1@10&F1@25&F1@50& Edit & Acc  & Avg \\    
\hline      
    
& \checkmark  & \checkmark  & \checkmark  & \checkmark  & \checkmark  & \checkmark  & \checkmark  & \checkmark  & \checkmark  &  83.6 & 83.6 & 72.8 & 78.5 & 84.3 & 80.6 \\
\checkmark  &             & \checkmark  & \checkmark  & \checkmark  & \checkmark  & \checkmark  & \checkmark  & \checkmark  & \checkmark  & 85.6 & 84.1 & 77.1 & 81.6 & 83.7 & 82.4 \\
\checkmark  & \checkmark  &          & \checkmark  & \checkmark  & \checkmark  & \checkmark  & \checkmark  & \checkmark  & \checkmark  & 85.2 & 83.9 & 76.9 & 80.7 & 83.9 & 82.1 \\
\checkmark  & \checkmark  & \checkmark  &          & \checkmark  & \checkmark  & \checkmark  & \checkmark  & \checkmark  & \checkmark  & 86.1 & 83.8 & 74.5 & 79.3 & 83.3 & 81.4 \\
\checkmark  & \checkmark  &          &         & \checkmark  & \checkmark  & \checkmark  & \checkmark  & \checkmark  & \checkmark  & 85.3 & 82.5 & 74.7 & 77.9 & 82.8 & 80.6 \\
 & \checkmark  &       &          & \checkmark  & \checkmark  & \checkmark  & \checkmark  &      \checkmark      & \checkmark  & 82.4 & 79.7 & 73.1 & 76.7 & 83.2 & 79.0 \\
 \checkmark  & \checkmark  &  \checkmark   &  \checkmark   &   \checkmark   &   & \checkmark  & \checkmark  & \checkmark  & \checkmark  & 82.8 & 80.5 & 71.6 & 76.1 & 82.6 & 78.7 \\
\checkmark  & \checkmark  &  \checkmark   &  \checkmark   &      & \checkmark  & \checkmark  & \checkmark  & \checkmark  & \checkmark  & 82.8 & 80.5 & 71.6 & 76.1 & 82.6 & 78.7 \\
\checkmark  & \checkmark  &  \checkmark   &  \checkmark   &   \checkmark   &  \checkmark  &   &   &    & \checkmark  & 84.3 & 82.1 & 74.4 & 78.8 & 83.5 & 80.6 \\
\checkmark  & \checkmark  & \checkmark  & \checkmark  & \checkmark  & \checkmark  & \checkmark  & \checkmark  & \checkmark  &  & 82.7 & 81.9 & 73.0 & 78.3 & 84.9 & 80.2 \\
\checkmark  & \checkmark  &  \checkmark   &  \checkmark   &   \checkmark   &  \checkmark  &   &   &    &   & 64.1 & 62.1 & 54.6 & 54.7 & 84.7 & 64.1 \\
\checkmark  & \checkmark  & \checkmark  & \checkmark  & \checkmark  & \checkmark  & \checkmark  &             & \checkmark  & \checkmark  & 87.5 & 85.6 & 79.4 & 81.1 & 85.1 & 83.7 \\
\checkmark  & \checkmark  & \checkmark  & \checkmark  & \checkmark  & \checkmark  & \checkmark  & \checkmark  &            & \checkmark  & 89.3 & 88.1 & 81.6 & 83.8 & 87.4  & 86.1\\
\checkmark  & \checkmark  & \checkmark  & \checkmark  & \checkmark  & \checkmark  & \checkmark  & \checkmark  & \checkmark  & \checkmark  & \textbf{89.5} & \textbf{88.3} & \textbf{82.5} & \textbf{83.9} & \textbf{88.0} & \textbf{86.4} \\

\hline  
\end{tabular} 
\label{table:abl}
\end{table*}

\subsection{Results and comparison}

\subsubsection{Supervised action segmentation}

Table \ref{table:sup} shows the comparison with existing methods on supervised action segmentation with a fixed smoothing window of 80, 40, and 50 for 50Salads, GTEA and Breakfast datasets, respectively. On the 50Salads dataset, ASESM demonstrates superior performance compared to state-of-the-art methods. Specifically, a 2\% accuracy improvement over LTContext \cite{bahrami2023much}, despite their use of local attention and sparse long-term context attention. On GTEA dataset, the evaluation on F1 score with boundary correction applied after the initial predictions during test time shows better performance than the other methods except for the accuracy, with a second better performance. The evaluation on Breakfast dataset shows higher accuracy and on par of the other methods on F1 and Edit. 

Boundary correction and segment smoothing can be applied as post-processing on existing models. In Table \ref{tab:asformer}, we demonstrate the effectiveness of both on improving the segmentation accuracy of ASFormer \cite{yi2021asformer}. To clearly validate the consistency in improving the performance, the evaluation on each split of 50Salads dataset is presented. The results show that the average performance is always improved when applying boundary correction, smoothing, and much more when both are included as post-processing, especially on F1 and Edit.

\subsubsection{Unsupervised action segmentation}

Table \ref{table:unsup} shows the comparison results of the unsupervised detection-correction algorithm with  existing unsupervised methods on Breakfast and 50Salads datasets.  On Breakfast dataset, with a boundary interval of $B\textsubscript{intrv}=300$, the results show superior performance compared to state-of-the-art methods with F1 score of 63.2. It surpassed  TSA \cite{bueno2023leveraging} , which based on metric learning, with a margin of 5.2.  While no learning is required in our detection-correction algorithm. On 50Salads, the performance is also better than the two compared methods UDE \cite{swetha2021unsupervised} and TOT \cite{kumar2022unsupervised}.

We further compare the unsupervised algorithm with a semi-supervised method SemiTAS \cite{ding2022leveraging} in Table \ref{table:semi}.  On 50Salads dataset with a fixed boundary interval $B\textsubscript{intrv}=500$, our algorithm achieves F1 score of 51.2 higher than SemiTAS in both cases, when they use  10\% or 5\% of labeled data. Using an automatic dynamic boundary interval threshold leads to F1 of 43.7, which is still better than SemiTAS using 5\% of labeled data. In the case of GTEA dataset with $B\textsubscript{intrv}=70$, the F1 score still shows better performance in case when they use 5\% of labels.

In Table \ref{table:timestamp}, we show the performance against two weakly supervised timestamps methods with both our supervised and unsupervised results. For unsupervised learning, given that they use training data with pseudo labels and we don't use any training data, our results are still acceptable, achieving accuracies of 70.4 and 63.2 on the GTEA and Breakfast datasets, respectively, though lower than the comparative accuracies of 80.9 and 72.0. However, for supervised learning, it is clear that our method performs better since we use full frame-wise labels.

\subsection{Ablation study}

\subsubsection{Effect of each component of the supervised framework}

The ablation study in Table \ref{table:abl} evaluates the contribution of each component within the supervised learning architecture by measuring F1 scores at different overlap thresholds, Edit distance, and accuracy on split 1 of the 50Salads dataset. The reference results, which include all components, achieve the highest performance with an average F1 score of 86.4. Specifically, the model attains F1@10, F1@25, and F1@50 scores of 89.5, 88.3, and 82.5, respectively, an Edit distance of 83.9, and an accuracy of 88.40 (last row). These results serve as the baseline for assessing the impact of omitting individual components.

The ablation study reveals that omitting individual encoders (rows 1 to 4) results in a decline in performance, with the average dropping to 80.6, 82.4, 82.1, and 81.4, respectively. When two or three encoders are removed (rows 5 and 6), the performance declines further to 80.6 and 79.0, respectively. This suggests that each encoder contributes to extracting distinct levels of features from the data, and their combined effect is crucial for achieving high performance. The voting stage, which considers the similarity of predictions from different encoders, plays a significant role in enhancing the model's robustness.

The study also examines the effect of removing the voting mechanism by replacing it with a frame-wise sum of the probability predictions generated by the four encoders. The results in row 7 show a severe decline in performance, with the average dropping to 78.7. Specifically, F1@10, F1@25, and F1@50 scores decrease to 82.8, 80.5, and 71.6, respectively. The Edit distance and accuracy also decline to 76.1 and 82.6. This highlights the importance of the voting mechanism in considering the similarity between predictions generated from global and local feature representations of the frames, which enhances the robustness and reliability of the final output. Additionally, the removal of the TCB block (row 8) leads to a decline in performance, with the average dropping to 78.7. This demonstrates that the TCB is essential for capturing temporal dependencies within the frame-wise prediction sequence.

\begin{table}
\footnotesize
\setlength{\tabcolsep}{3.5pt}
\renewcommand{\arraystretch}{1.3}
\caption{Effect of different sizes of the smoothing windows S\textsubscript{win} on the performance for split 1 of 50Salads and GTEA datasets.}
\centering
\begin{tabular}{l l l l l l l }
\hline
               &   & 50Salads 	&	&	&	&	\\
          
S\textsubscript{win}&	F1@10&F1@25&F1@50	&		Edit&	Acc &	Avg\\
\hline
           60 &	86.7    &	85.2	&   78.2&	81.0&	84.3 & 83.1\\
           70 &	84.2    &	82.8    &	76.5&	77.4&	84.7 & 81.1\\
           80 &	87.7    &   86.3    &   80.2 & 82.4 & 85.3 & 84.4\\
           MaxDistBound/10 &  84.9     &   83.0 & 74.5&	  76.9&	  84.0 & 80.7\\

         \hline 
               &   & GTEA 	&	&	&	&	\\
                
S\textsubscript{win} &	F1@10&F1@25&F1@50	&	        Edit&	Acc &	Avg\\
\hline
            4&	89.1&	88.4&	80.4&	85.7&	81.8& 85.1\\
            10&	88.2	&86.0	&79.4&	84.5&	78.5& 83.3\\
            20&	87.9	&83.3&	72.0&	84.2&	79.5&81.4\\
          MaxDistBound/10&  87.9 & 86.4 & 75.0	&  82.8& 79.7&82.7\\

\hline
\end{tabular}
\label{table:smooth}
\end{table}

\begin{table}
\footnotesize
\renewcommand{\arraystretch}{1.3}
\caption{Effect of the boundary interval threshold (B\textsubscript{intrv}) on the unsupervised action segmentation algorithm on the three datasets.}
\centering
\begin{tabular}{l l |l l| l l }
\hline
50Salads& & GTEA & &  Breakfast&   \\
B\textsubscript{intrv} & F1 & B\textsubscript{intrv} & F1 & B\textsubscript{intrv} & F1 \\
\hline
200 & 36.9   &  40 & 65.1& 270  &61.9 \\
300 & 44.9    & 50 & 68.4 &280 & 61.5\\
400 & 49.3   &  60 & 67.2 &290 & 62.7\\
500&  51.2   &  70 & 70.4 &300 & 63.2\\
600 & 49.2   &  80 & 65.8 &310 & 62.7\\
700  &47.0   &  90 & 63.2 &320 &62.8\\
800&  47.2   &  100& 57.8 &330 & 62.6\\
\hline
\end{tabular}
\label{table:intrv}
\end{table}

\begin{table}
\footnotesize
\setlength{\tabcolsep}{1.3pt}
\renewcommand{\arraystretch}{1}
\centering
\caption{\blue{Improvement over ASFormer \cite{yi2021asformer} using boundary correction and smoothing as post-processing .}}
\begin{tabular}{@{}LLLLLLLLL@{}}
\toprule
Split & Config & F1@10 & F1@25 & F1@50 & Edit & Acc & Avg \\ \midrule
\multirow{4}{*}{Sp1} & ASFormer      & 82.1 & 80.8 & 72.4 & 76.4 & 82.9  & 78.9  \\
                        & ASFormer + BC        & \textbf{83.6} & \textbf{81.8} & \textbf{74.2} & \textbf{77.7} & 82.4 & \textbf{79.9}\\
                        & ASFormer + Smth     & \textbf{85.5} & \textbf{84.0} & \textbf{73.2} & \textbf{80.6} & 81.4 & 	\textbf{80.9}\\
                        & ASFormer + BC+ Smth   & \textbf{86.1} & \textbf{84.6} & \textbf{73.8} & \textbf{80.9} & 81.5 & \textbf{81.4}\\ \midrule
\multirow{4}{*}{Sp2} & ASFormer      & 86.1 & 84.6 & 77.9 & 80.4 & 87.8 & 83.4\\
                        & ASFormer + BC        & 86.1 & 84.6 & 77.9 & 80.4 & 87.4& 83.3\\
                        & ASFormer + Smth     & \textbf{87.8} & \textbf{86.8} & 77.8 & \textbf{81.4} & 86.8& \textbf{84.1}\\
                        & ASFormer + BC+ Smth  & \textbf{88.3} & \textbf{87.3} & \textbf{78.9} & \textbf{82.3} & 87.4 & \textbf{84.8}\\ \midrule
\multirow{4}{*}{Sp3} & ASFormer      & 86.3 & 85.9 & 76.4 & 82.1 & 86.1 & 83.4\\
                        & ASFormer + BC        & \textbf{86.5} & \textbf{86.1} & 76.1 & \textbf{82.4} & 85.6 & 83.3\\
                        & ASFormer + Smth     & \textbf{86.6} & \textbf{86.1} & \textbf{78.2} & \textbf{83.2} & 84.7 &\textbf{83.8}\\
                        & ASFormer + BC+ Smth   & \textbf{87.9} & \textbf{87.9} & \textbf{78.9} & \textbf{83.5} & 85.0 & \textbf{84.6}\\ \midrule
\multirow{4}{*}{Sp4} & ASFormer      & 83.7 & 80.8 & 75.0 & 79.0 & 82.6 &80.2\\
                        & ASFormer + BC        & \textbf{83.6} & \textbf{81.2} & 74.9 & 78.6 & 82.5 &80.2\\
                        & ASFormer + Smth      & \textbf{85.2} & \textbf{82.7} & 74.7 & 78.4 & 81.7 & \textbf{80.5}\\
                        & ASFormer + BC+ Smth   & \textbf{84.9} & \textbf{82.4} & \textbf{75.4} & 77.8 & 82.2&\textbf{80.5} \\ \midrule
\multirow{4}{*}{Sp5} & ASFormer      & 87.3 & 84.8 & 78.3 & 78.0 & 88.7 &83.4\\
                        & ASFormer + BC        & 87.3 & 85.8 & 78.3 & 80.0 & 88.2& \textbf{83.9} \\
                        & ASFormer + Smth      & \textbf{89.6} & \textbf{87.6} & \textbf{80.8} & \textbf{81.9} & 87.4& \textbf{85.5}\\
                        & ASFormer + BC+ Smth   & \textbf{89.6} & \textbf{87.6} & \textbf{80.8} & \textbf{81.9} & 87.5& \textbf{85.5} \\ \midrule
\multirow{4}{*}{Avg} & ASFormer      & 85.1 & 83.4 & 76.0 & 79.6 & 85.6& 81.9\\
                        & ASFormer + BC   & \textbf{85.4} & \textbf{83.9} & \textbf{76.3} & \textbf{79.8} & 85.2& \textbf{82.1}\\
                        & ASFormer + Smth & \textbf{86.9} & \textbf{85.5} & \textbf{77.0} & \textbf{81.1} & 84.4& \textbf{83.0}\\
                        & ASFormer + BC+ Smth & \textbf{87.4} & \textbf{86.0} & \textbf{77.6} & \textbf{81.3} & 84.7 &\textbf{83.4}\\ \bottomrule

\end{tabular}
\label{tab:asformer}
\end{table}

When all components are present except for boundary correction after the voting stage at test time (row 12), the performance decline is relatively small, with the average F1 score dropping to 83.7. However, ignoring boundary correction in the post-processing step (row 13) has a minimal impact on performance. This observation suggests that boundary correction has a more significant impact during training, where it is learned over multiple iterations, compared to its application during the single-pass inference phase at test time.

Omitting boundary correction from the framework in both training  and testing leads to a decline in the performance to an average of 80.6. Specifically, F1@10, F1@25, and F1@50 scores decrease to 84.3, 82.1, and 74.4, respectively, while the Edit distance and accuracy drop to 78.8 and 83.5. These results highlight the significant impact of incorporating feature similarity with initial boundary predictions to enhance segmentation accuracy through the boundary correction algorithm.

The absence of smoothing in the post-processing phase has a similar impact to the absence of boundary correction, with a decline to an average of 80.2. However, omitting both boundary correction and smoothing from the framework leads to the lowest performance among all the analysis cases with an average of 64.1. These results show the impact of the contribution of our work on the learning scheme and validate the initial assumption that explicit similarity measurement enhances segmentation accuracy.

\begin{figure*}
\centering
\includegraphics[width=1\linewidth]{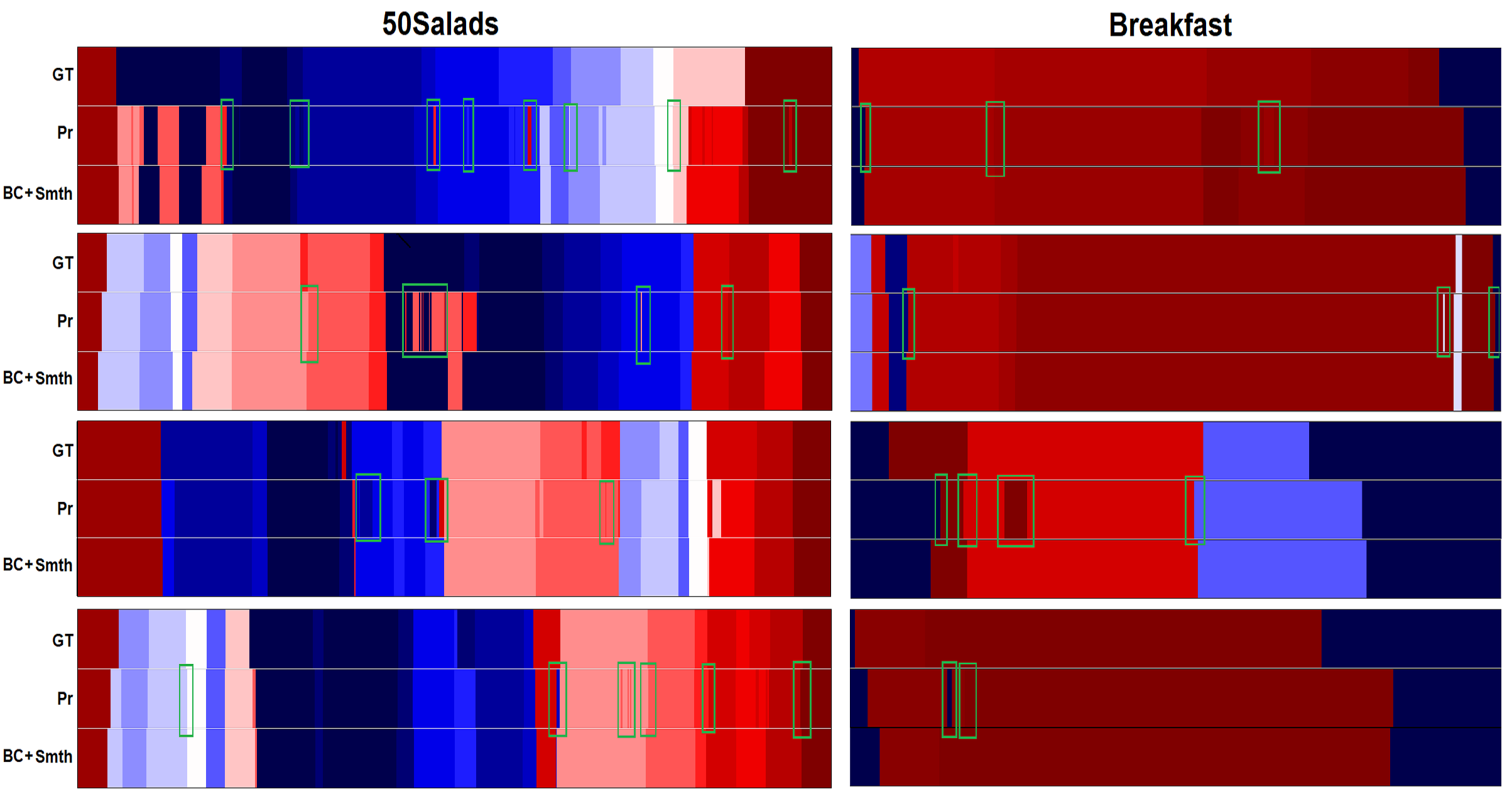}
\caption{Qualitative results of supervised action segmentation on testing videos from split 1 of 50Salads dataset (left), and from split 2 of Breakfast dataset (right). GT: Ground truth, Pr: Initial prediction, BC+Smth: Applying boundary correction and smoothing. Corrected predictions are marked with green rectangles.}
\label{fig:vis} 
\end{figure*}

For further evaluation of our contribution, in Fig. \ref{fig:vis}, we show qualitative action segmentation results of our supervised approach in the presence and the absence of boundary correction and smoothing. 

\subsubsection{Effect of the smoothing window size S\textsubscript{win}}

Table \ref{table:smooth} shows the impact of different smoothing window sizes (S\textsubscript{win}) on performance for split 1 of the 50Salads and GTEA datasets. On 50Salads, S\textsubscript{win}=80 achieves the highest average score of 84.4. Decreasing the window size to 60 and 70 results in performance drops to 83.1 and 81.1, respectively. Increasing the window beyond 80 also reduces performance. Using an automatically generated window yields a lower average score of 80.7. On GTEA, a small S\textsubscript{win}=4 achieves a performance of 85.1, while the automatic window performs less than a fixed window.

From this analysis, we conclude that longer videos with more actions require larger smoothing windows. For example, 50Salads videos (\textasciitilde9000 frames, 20 actions) perform best with S\textsubscript{win}=80, while GTEA videos (\textasciitilde1000 frames, 11 actions) achieve optimal results with S\textsubscript{win}=4.

\begin{table}
\setlength{\tabcolsep}{3.5pt}
\renewcommand{\arraystretch}{1.4}
\footnotesize
\caption{Comparison between ASESM and ASFormer on the number of parameters FLOPs for split 1 of 50Salads dataset.}
\centering
\begin{tabular}{l l l l l l }
\hline
Method&	Param(M)& \blue{FLOPs(G)}& Avg\\ 

\hline     

            ASFormer \cite{yi2021asformer}	&  1.13 & \blue{4.79} &79.3\\
           \textbf{ASESM} (E\textsubscript{2}+TCB) & 1.29& \blue{4.2}	&79.0\\
           \textbf{ASESM} (E\textsubscript{1}+E\textsubscript{2}+TCB) & 1.44& \blue{4.31}	& 80.6\\
           \textbf{ASESM} (E\textsubscript{1}+E\textsubscript{2}+E\textsubscript{3}+TCB)	& 2.99& \blue{5.09}	& 81.4\\
           \textbf{ASESM} (E\textsubscript{1}+E\textsubscript{2}+E\textsubscript{3}+E\textsubscript{4}+TCB) & 8.64& \blue{7.55}	& 84.4 \\

\hline
\end{tabular}
\label{table:comput}
\end{table}

\subsubsection{Effect of the boundary interval threshold B\textsubscript{intrv} on unsupervised segmentation}

Table \ref{table:intrv} explores the impact of the boundary interval threshold (B\textsubscript{intrv}) in  the unsupervised detection-correction algorithm on the three datasets. On 50Salads dataset,  when there is an increase of B\textsubscript{intrv} from 200 to 800, the performance reaches its peak with 500. On GTEA, the optimal threshold is 70 within an interval of 40 to 100. For Breakfast, the best threshold is 300 when tested between 270 and 330. Generally, the performance gradually declines as the interval moves away from the optimal value, either lower or higher. It is highly likely that the optimal interval depends on video length and the number of actions. However, using an automatic interval still yields competitive performance.

\subsection{Computation complexity}

Table \ref{table:comput} explores how ASESM scales in both parameter complexity and performance compared to ASFormer \cite{yi2021asformer} on split 1 of 50Salads dataset. ASFormer requires 1.13 million parameters with 4.79 FLOPs and achieves an average performance of 79.3. ASESM with just one encoder (ENC\textsubscript{2}) and TCB require a 1.29m parameter with 4.2 FLOPs, with a slightly lower performance than ASFormer. Increasing complexity by adding ENC\textsubscript{1} alongside ENC\textsubscript{2} and TCB raises the parameter count to 1.44 million with 4.31 FLOPs, yielding an improved average performance of 80.6. Incorporating ENC\textsubscript{1}, ENC\textsubscript{2}, ENC\textsubscript{3}, and TCB results in 2.99 million parameters with 5.09 FLOPs and enhances the average performance to 81.4. The most complex ASESM configuration, with all encoders and TCB, uses 8.64 million parameters and 7.55 FLOPs, achieving the highest average performance of 83.5.

\section{Conclusion}  \label{section5}
We introduced ASESM (Action Segmentation via Explicit Similarity Measurement), a new approach that enhances performance by incorporating explicit similarity evaluation into the learning process. Our supervised learning architecture leverages multi-resolution frame-level features processed through multiple transformer encoders for precise initial predictions via frame-wise similarity voting. We also proposed a boundary correction algorithm that refines predictions by analyzing feature similarity between consecutive frames, followed by multi-stage refinement using temporal convolutions. Additionally, our method includes a post-processing phase with optional boundary correction and segment smoothing to eliminate outlier classes. Furthermore, we proposed an unsupervised boundary detection-correction algorithm that identifies segment boundaries based solely on feature similarity, without the need for training. The experimental results and the ablation study demonstrate the effectiveness of our method. Moreover, the proposed boundary correction and smoothing enhance the segmentation performance of other backbones when applied as post-processing.

Although the method demonstrates better performance in both supervised and unsupervised action segmentation, it involves several independent steps, particularly in the supervised framework. This complexity can lead to adjusting many parameters.  As a future work, we aim to develop a unified unsupervised algorithm that integrates supervised boundary correction, boundary detection-correction and smoothing with an unsupervised representation learning as an initial step, for a fully single unsupervised framework.






\section*{Acknowledgement}

Kamel Aouaidjia and Chongsheng Zhang are supported
in part by the National Natural Science Foundation of
China (No.62250410371) and the Henan Provincial Key R\&D
Project (No.232102211021). 









\bibliographystyle{cas-model2-names}

\bibliography{cas-refs}

\end{document}